\pdfoutput=1

\documentclass[11pt]{article}

\usepackage[final]{mtsummit25}
\usepackage{copyright}

\usepackage{times}
\usepackage{latexsym}

\usepackage[T1]{fontenc}

\usepackage[utf8]{inputenc}

\usepackage{microtype}

\usepackage{inconsolata}

\usepackage{graphicx}

\title{Are AI agents the new machine translation frontier? Challenges and opportunities of single- and multi-agent systems for multilingual digital communication}

\author{Vicent Briva-Iglesias \\
SALIS, CTTS, ADAPT Centre \\
Dublin City University \\
\texttt{vicent.brivaiglesias@dcu.ie}
}

\begin{document}
\maketitle
\begin{abstract}
The rapid evolution of artificial intelligence (AI) has introduced AI agents as a disruptive paradigm across various industries, yet their application in machine translation (MT) remains underexplored. This paper describes and analyses the potential of single- and multi-agent systems for MT, reflecting on how they could enhance multilingual digital communication. While single-agent systems are well-suited for simpler translation tasks, multi-agent systems, which involve multiple specialized AI agents collaborating in a structured manner, may offer a promising solution for complex scenarios requiring high accuracy, domain-specific knowledge, and contextual awareness. To demonstrate the feasibility of multi-agent workflows in MT, we are conducting a pilot study in legal MT. The study employs a multi-agent system involving four specialized AI agents for (i) translation, (ii) adequacy review, (iii) fluency review, and (iv) final editing. Our findings suggest that multi-agent systems may have the potential to significantly improve domain-adaptability and contextual awareness, with superior translation quality to traditional MT or single-agent systems. This paper also sets the stage for future research into multi-agent applications in MT, integration into professional translation workflows, and shares a demo of the system analyzed in the paper.
\end{abstract}

\section{Introduction}
In an increasingly interconnected world, the demand for accurate, efficient, and context-aware multilingual communication has surged, driven by globalization and digital transformation \citep{zahidiFutureJobsReport2025}. MT systems face persistent challenges in handling domain-specific jargon, adapting to contextual particularities, and aligning with client-specific guidelines \citep{kennyHumanMachineTranslation2022}. Traditional neural machine translation (NMT) models, though advanced, often operate as monolithic systems, lacking the flexibility to dynamically integrate specialized knowledge or iterative quality controls without fine-tuning, a critical problem in high-stakes domains such as legal, medical, or technical translation \citep{briva-iglesiasTraduccionHumanaVs2021,montalt-resurreccioPatientCentredTranslationCommunication2024}.

\begin{figure}[t]
  \includegraphics[width=\columnwidth]{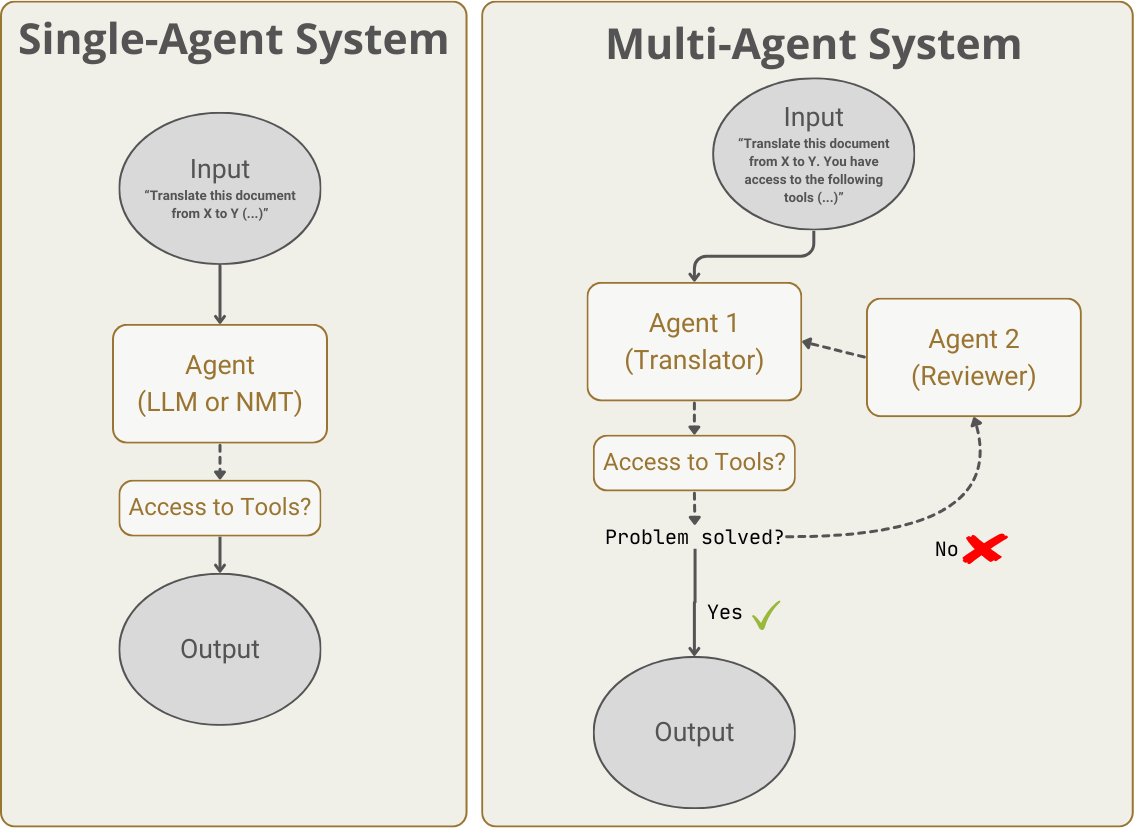}
  \caption{An example of single- and multi-agent systems applied to MT.}
  \label{fig:single-multi}
\end{figure}

The emergence of AI agents—autonomous or semi-autonomous systems capable of reasoning about tasks, tool integration, and taking actions to achieve specific goals—may present a paradigm shift for MT. Increasingly adopted in fields like software engineering \citep{qianChatDevCommunicativeAgents2024}, customer support \citep{liMetaAgentsSimulatingInteractions2023}, data analysis \citep{wangAlphaGPTHumanAIInteractive2023}, and academic research \citep{schmidgallAgentLaboratoryUsing2025}, AI agents remain underexplored in translation workflows. In the context of MT, AI agents can be organized into single-agent systems for straightforward tasks or multi-agent systems for complex workflows requiring collaboration and iterative refinement (see Figure \ref{fig:single-multi}). By leveraging highly customisable workflows, external tools (e.g., domain-specific glossaries, translation memories), memory, and advanced planning capabilities, multi-agent systems may be able to address the limitations of traditional MT systems. For instance, by decomposing translation tasks into specialized roles (e.g., translation, adequacy review, fluency editing) and enabling dynamic interaction between AI agents, multi-agent systems may mirror professional human workflows. 

The primary goal of this paper is to explore the capabilities of AI agent workflows for MT, with a focus on their organization, customization, and generalisability across fields requiring multilingual digital communication. This paper investigates the following research questions (RQs): 

\textbf{RQ1.} \textit{How effective are multi-agent systems in legal MT compared to single-agent approaches?}

\textbf{RQ2.} \textit{How do AI agent-based workflows align with professional human translation processes?}

\textbf{RQ3.} \textit{How does model temperature impact translation performance in multi-agent systems?}

\textbf{RQ4.} \textit{How does model size impact translation performance in multi-agent systems?}

This work makes several key contributions to the MT field. Sections \ref{sec:2} and \ref{sec:3} provide a theoretical framework for organizing AI agents into single-agent and multi-agent systems, highlighting the use of customizable workflows and external tools. Section \ref{sec:4} analyses the practical application of this framework through a pilot study, illustrating the potential of AI agents in translation workflows. The pilot study consists of four specialized agents for legal translation: (i) a Translator-Agent, (ii) an Adequacy Reviewer-Agent, (iii) a Fluency Reviewer-Agent, and (iv) an Editor-Agent. This structure simulates real-world translation processes in legal settings, where consistency, terminology accuracy, and compliance are paramount. 

While the pilot study provides a practical example of AI agents in action, the broader focus of this paper is on the theoretical and methodological implications of AI agent workflows for MT, emphasizing their potential to transform multilingual communication across diverse fields, offering a foundation for future research and implementation. We also share a multi-agent public demo for further analysis and replication: \href{https://agents-parallel-2.streamlit.app/}{https://agents-parallel-2.streamlit.app/}.

\section{What are AI agents?}
\label{sec:2}
The concept of AI agents traces its roots to early AI research, where "rational agents" were defined as entities capable of autonomous action in pursuit of objectives \citep{russell1995intelligent}. However, until very recently, most agent systems relied on rigid algorithmic structures \citep{mnihHumanlevelControlDeep2015,lillicrapContinuousControlDeep2019}. The emergence of large language models (LLMs) has marked a significant turning point in AI agent systems, with enhanced reasoning and contextual understanding capabilities, allowing for more flexible and adaptable workflows \citep{brownLanguageModelsAre2020}. This has transformed AI agents from theoretical constructs into practical tools \citep{wangSurveyLargeLanguage2024}. We could now define AI agents as autonomous or semi-autonomous software programs designed to reason about tasks and execute actions to achieve predefined goals. Unlike traditional MT systems, which operate as static pipelines where an input in the source language is received by the system and an output in the target language is generated, agents can dynamically adapt their behaviour through a defined set of instructions, which allow them to plan, integrate tools, and iteratively refine their output \citep{chengExploringLargeLanguage2024}. All these advancements have facilitated the emergence of structured AI agent workflows, which can broadly be categorized into single-agent workflows and multi-agent workflows.
 
Single-agent workflows involve only one AI agent that performs tasks within a given environment. These agents function independently, performing sequential tasks such as summarizing, translating, or processing data. They often rely on predefined prompts and reinforcement mechanisms to enhance performance \citep{chengExploringLargeLanguage2024}. For example, in software engineering, single-agent systems have been successfully applied to automated debugging and code generation \citep{kimLanguageModelsCan2023}. In MT, a single-agent workflow could be instructing a traditional NMT system to translate something from an API call and/or using an LLM with a simple prompt for MT. Substantial research on the topic has already been conducted \citep{hendyHowGoodAre2023, gaoHowDesignTranslation2023, briva-iglesiasLargeLanguageModels2024a}.

Multi-agent workflows consist of different AI agents collaborating to achieve a shared objective. These workflows enable specialization, with each AI agent performing a designated role within a sequential or iterative system \citep{huDecentralizedClusterFormation2021}. Multi-agent workflows have seen widespread adoption in domains such as software engineering (e.g., GitHub Copilot for code generation) \citep{qianChatDevCommunicativeAgents2024},  customer service (e.g., chatbots for query resolution) \citep{liMetaAgentsSimulatingInteractions2023}, data analysis (e.g., automated report generation) \citep{wangAlphaGPTHumanAIInteractive2023} or academic research \citep{schmidgallAgentLaboratoryUsing2025}. Their success in these fields stems from their ability to decompose tasks into subtasks, collaborate with external tools (web search, specific databases, etc.), and optimize outcomes through feedback loops, memory and/or reasoning. Multi-agent systems have become a focal point of AI research due to their ability to tackle complex problems requiring distributed decision-making and contextual adaptation \citep{zhugeMindstormsNaturalLanguageBased2023}. Several studies highlight the advantages of multi-agent collaboration, particularly in tasks requiring high levels of reasoning and iterative improvement \citep{gurRealWorldWebAgentPlanning2024, dongSelfcollaborationCodeGeneration2024}. For instance, research on AI planning and task execution has demonstrated that multi-agent workflows lead to improved adequacy and efficiency compared to single-agent approaches \citep{schmidgallAgentLaboratoryUsing2025}.

However, the application of AI agents in MT remains scarce, despite the alignment between agent-based workflows and the iterative, role-driven nature of professional translation processes. While traditional MT research prioritized model architecture improvements \citep{vaswaniAttentionAllYou2023}, the integration of AI agent workflows—inspired by frameworks like ReAct \citep{yaoReActSynergizingReasoning2023} and multi-step planning—represents a shift toward mimicking human translation teams’ collaborative dynamics. To date, only a few experiments on AI agents for MT have been published. For instance, \citet{wuPerhapsHumanTranslation2024} introduced TransAgents, a multi-agent system designed to translate ultra-long literary texts. This system mimicked human editorial workflows by incorporating specialized agents for different translation tasks, including initial translation, localization, proofreading, and quality assessment. The authors report that despite achieving lower d-BLEU scores, TransAgents-generated translations were preferred by human evaluators over conventional MT systems and even human references due to improved cultural and contextual adaptation. It is worth stressing, however, that the MT evaluation was not conducted by professional evaluators and could therefore have had an impact on the results \citep{laubliSetRecommendationsAssessing2020}. \citet{ngAndrewyngTranslationagent2025} introduced another multi-agent workflow for MT, using three different AI agents: the first agent translates a text, the second agent provides improvement suggestions, a third agent produces a final translation after considering the suggestions. The author reports using BLEU on standard translation datasets and suggests that this workflow has shown mixed results—sometimes competitive with, and occasionally falling short of, leading commercial translation systems—but no specific details nor human evaluation have been found. More recently, \citet{sinSolvingUnsolvableTranslating2025} proposed a multi-agent system for translating Hong Kong legal judgments, comprising Translator, Annotator, and Proofreader agents powered by GPT-3.5 Turbo, and a memory-based few-shot prompting strategy was used for iterative quality improvement. The evaluation shows that the multi-agent system outperformed both traditional MT systems and even GPT-4o in accuracy, coherence, and style, offering a scalable solution for bilingual legal translation. This demonstrates that multi-agent systems for MT are a nascent area of research with great potential for further enhancement that lacks further empirical analysis.

\section{The potential of AI agents for MT}
\label{sec:3}
From our perspective, the efficacy of AI agents in MT depends on four core attributes:

\textbf{Autonomy:} AI agents operate independently or with minimal human oversight once configured, provided they receive clear instructions (e.g., roles and tasks to conduct, style preferences, domain constraints). For instance, a Translator-Agent in a legal translation workflow could be instructed to provide translations while adhering to jurisdictional terminology from a specific country.

\textbf{Tool use:} Agents can integrate external resources such as translation memory systems, domain-specific databases (e.g., legal glossaries from the client), and retrieval-augmented generation (RAG) frameworks to enhance accuracy and consistency \citep{lewisRetrievalAugmentedGenerationKnowledgeIntensive2020}. Early works have demonstrated the promising results of RAG for MT \citep{liSurveyRetrievalAugmentedText2022, coniaCrossCulturalMachineTranslation2024}. For example, the above Translator-Agent could cross-reference terminology from previously translated materials from a specific client to ensure compliance or have access to IATE, if working with legal documents.

\textbf{Memory:} Agents can learn from feedback loops, refining outputs iteratively \citep{mnihHumanlevelControlDeep2015}. For example, a Fluency Reviewer-Agent might prioritize syntax and style corrections based on recurring errors flagged in prior iterations.

\begin{figure*}[t]
  \centering
  \includegraphics[width=\textwidth]{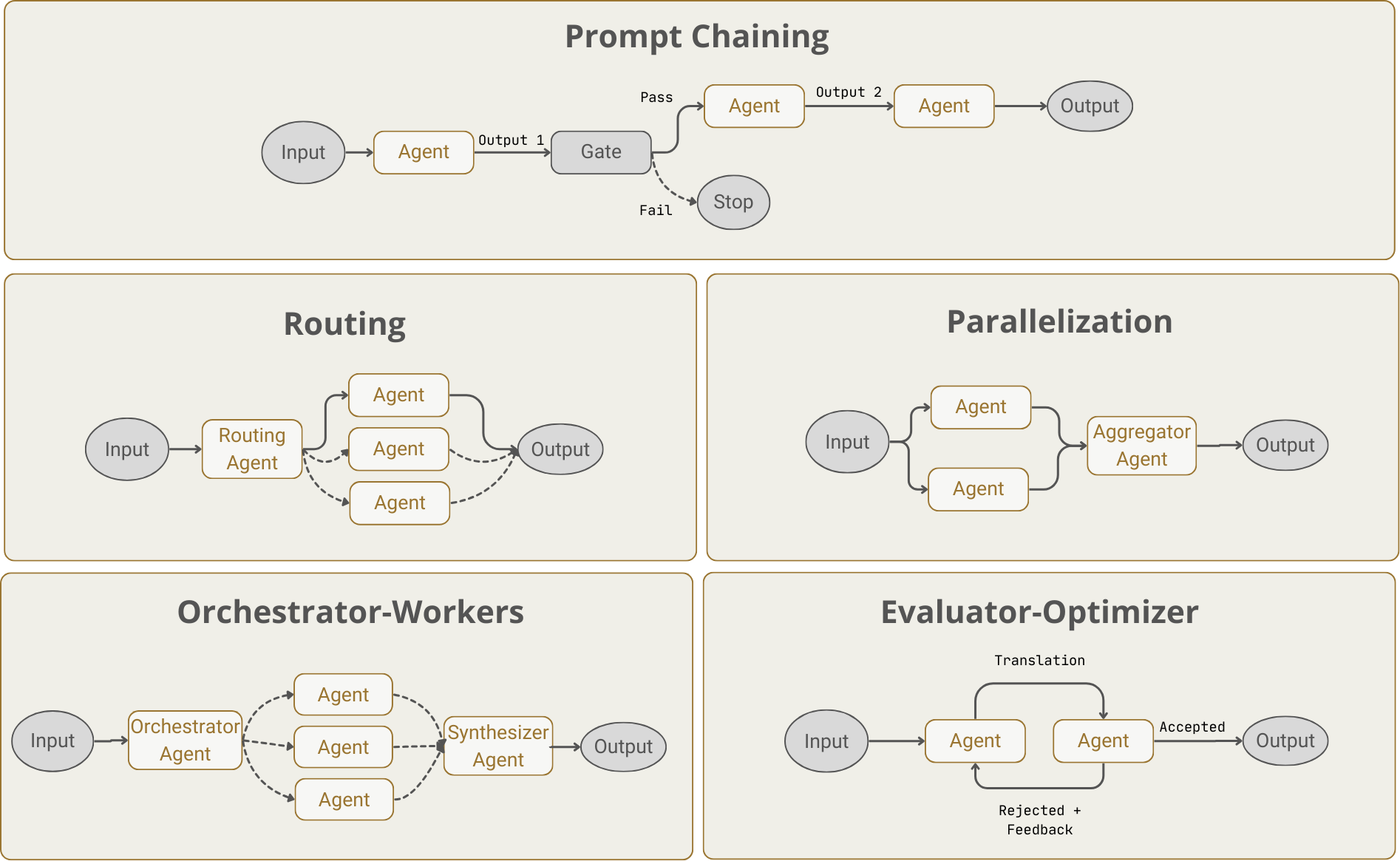}
  \caption{Some potential customisations of multi-agent workflows.}
  \label{fig:workflows}
\end{figure*}

\textbf{Workflow customization:} AI agents enable dynamic MT workflows through customizable architectures. Figure \ref{fig:workflows} depicts five potential multi-agent workflows (not exclusive) that we define considering their application to MT challenges such as domain adaptation, scalability, and quality assurance\footnote{Based on Anthropic's blog entry: https://www.anthropic.com/engineering/building-effective-agents}. Workflows can be sequential or iterative. A sequential AI agent workflow is a structured process where tasks are executed in a strict order, with each step depending on the completion of the previous one. An iterative AI agent workflow is more dynamic and allows multiple tasks to be performed simultaneously, with results being refined through back-and-forth adjustments.

\subsection{Prompt Chaining}
Prompt chaining is a structured, sequential workflow in which each step’s output serves as the input for the next, ensuring systematic reasoning and iterative refinement. In MT, this workflow may mirror professional translation processes by breaking tasks into specialized stages, allowing for greater control over adequacy, domain adaptation, and linguistic coherence. The process may begin with a preprocessing agent that analyses the source text, extracting relevant metadata such as document type, target audience, and domain-specific terminology. This preprocessing agent may leverage RAG or TM systems to enhance contextual precision. Next, a translation agent generates an initial draft by incorporating the retrieved information and applying domain-specific constraints to maintain terminological and syntactic adequacy. Finally, an automatic post-editing agent refines the translation, improving fluency, ensuring stylistic coherence, and verifying adherence to formatting or regulatory guidelines. By structuring translation tasks into interdependent steps, prompt chaining may improve quality control, enhance domain adaptability, and reduce errors.

\subsection{Routing}
Routing is an iterative workflow that could allocate translation tasks to specialized AI agents based on specific input characteristics, such as language pair, domain, or text complexity. By intelligently distributing tasks, this approach may optimize efficiency and ensure that each translation request is handled by the most suitable agent. In MT, multi-agent routing workflows may improve adaptability by directing different types of texts to agents equipped with the necessary linguistic and contextual expertise. For instance, low-resource languages, which often lack large-scale training data, can be assigned to agents fine-tuned on regional corpora to improve translation quality. Similarly, domain-specific texts such as legal contracts or medical reports can be routed to agents with access to specialized databases like legal termbases or medical corpora, ensuring compliance with industry standards and terminology consistency.

Beyond language and domain specialization, routing may also account for the complexity of the translation task. Simple texts can be processed using smaller models optimized for speed and efficiency, provided that quality can be lower and that the aim of the translation is of assimilation exclusively \citep{kennyHumanMachineTranslation2022}. In contrast, complex documents where dissemination is required, such as legal contracts or regulatory filings, may require a multi-agent review powered by bigger and better language models, where separate agents handle terminology validation, fluency refinement, and formatting compliance. By leveraging adaptive routing, multi-agent workflows may optimize processing efficiency, improve translation quality across diverse domains, and enable greater scalability in multilingual digital communication processes.

\subsection{Parallelization}
Parallelization is a workflow strategy that may enable the simultaneous execution of independent translation subtasks across multiple AI agents, significantly reducing processing time and enhancing scalability. Unlike sequential workflows, where each step builds upon the previous one, in a parallelization workflow, tasks can be distributed among specialized agents that work concurrently, with their outputs later aggregated into a cohesive final translation. In MT, this approach may be particularly beneficial for large-scale multilingual projects, where a single document needs to be translated into multiple languages simultaneously. For instance, separate AI agents can translate a technical report into Spanish, French, and Chinese at the same time, each using language-specific instructions. This method may optimize efficiency without compromising linguistic or terminological precision.

Parallelization may also enhance MT workflows through sectional processing and multitasking. A long-form document, such as a research paper or a legal contract, can be divided into sections or chapters, with different agents handling translation and summarization in parallel. Additionally, quality assurance tasks—such as adequacy verification, fluency enhancement, and bias detection—may be conducted concurrently by dedicated agents to improve overall translation quality. A practical example of this approach can be seen in e-commerce localization, where product descriptions may need to be translated into multiple languages while maintaining brand consistency. In such cases, separate AI agents handle English-to-Spanish, English-to-French, and English-to-Chinese translation tasks simultaneously, while an aggregator agent ensures uniform terminology and adherence to brand style guidelines.

\subsection{Orchestrator-Workers} 
The orchestrator-workers workflow is a sequential MT approach in which a central orchestrator agent decomposes a translation task into subtasks, delegates them to specialized worker agents, and synthesizes the results into a cohesive final output. This structure may mimic human translation team dynamics, where project managers distribute workload among translators and reviewers to ensure quality and consistency. By enabling scalable handling of complex documents, this workflow may enhance translation efficiency while maintaining domain-specific adequacy and linguistic coherence.

A potential application of the orchestrator-workers workflow may be in legal MT. In this scenario, the orchestrator agent first segments a legal document, such as a contract, into discrete clauses and assigns them to translator agents specializing in legal terminology. Once the initial translations are completed, worker agents handle specific quality assurance tasks: an Adequacy Reviewer-Agent validates terminology against jurisdiction-specific legal databases, while a Fluency Reviewer-Agent ensures syntactic clarity and readability. Finally, an Editor-Agent synthesizes all outputs, ensuring consistency in phrasing, formatting, and cross-clause references. This structured delegation allows for greater adequacy and quality control compared to monolithic translation models, making it particularly suited for high-stakes domains such as law, medicine, and finance, where document integrity is paramount.
 
\subsection{Evaluator-Optimizer}
The Evaluator-Optimizer workflow is an iterative refinement process in which MT outputs undergo systematic evaluation and optimization until they meet predefined quality standards. This approach may be particularly valuable for high-stakes domains such as legal, medical, and technical translation, where even minor inaccuracies can lead to serious consequences. Unlike traditional MT workflows that produce static outputs, this workflow may introduce continuous quality control through feedback loops, ensuring precision, domain adherence, and linguistic coherence. However, the problem in this workflow may lie in determining when to stop the feedback loop and in instructing the model to stop editing the output once a established set of criteria is met.

One potential example may be as follows: a Generator-Agent produces an initial translation, drawing from domain-specific resources such as legal corpora or medical guidelines. Next, an Evaluator-Agent assesses the translation for errors, checking terminology, compliance, and contextual adequacy. This agent flags inconsistencies, mistranslations, or ambiguous phrasing using specialized databases, such as jurisdictional termbases for legal texts or the World Health Organization databases for medical terminology. An Optimizer-Agent then refines the flagged sections, making necessary adjustments and reprocessing the text until the evaluator confirms that all quality requirements have been met. For instance, in medical translation, an evaluator agent might verify that drug names and dosages align with regulatory standards, prompting the optimizer agent to correct any discrepancies before finalizing the output. By implementing this cycle of evaluation and optimization, the workflow may significantly enhance translation reliability, making it well-suited for fields where adequacy and compliance are non-negotiable.

\section{The pilot study}
\label{sec:4}

\begin{figure*}[t]
  \centering
  \includegraphics[width=\textwidth]{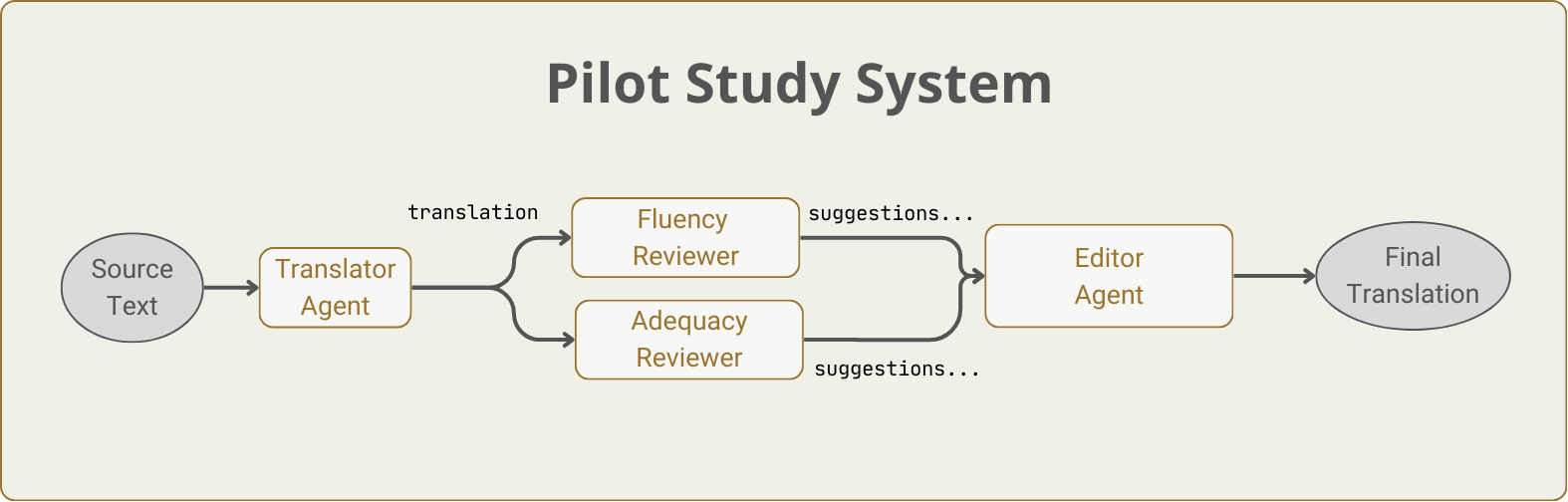}
  \caption{Multi-agent workflow analysed in the pilot study + demo.}
  \label{fig:demosystem}
\end{figure*}

The above workflows remain, at this stage, theoretical constructs designed to explore the potential of multi-agent systems in MT. While they provide a structured reflection on how AI agents could be leveraged to improve translation workflows, their practical feasibility, efficiency, and effectiveness in real-world applications have yet to be empirically validated. To bridge this gap, we are conducting a pilot study to assess the viability of AI agent-based approaches in professional translation settings. 

\subsection{The multi-agent workflow}
There is a growing number of libraries facilitating AI agent development. For this study, we employ LangGraph\footnote{Link to LangGraph: https://github.com/langchain-ai/langgraph} to construct a multi-agent system designed to simulate professional legal translation workflows. The system is built on a Parallelization workflow that integrates four specialized AI agents, each assuming a role that mirrors the functions of human legal translators and reviewers in an international organisation (see Figure \ref{fig:demosystem}). The agents operate in parallel, optimizing processing time while maintaining domain-specific quality controls.
 
The system’s workflow consists of the following AI agents. First, a Translator-Agent that produces an initial translation using an LLM. Even if we could have provided tool access to RAG and/or domain-specific databases, we only used the LLM as the information context. Second, we have two agents working in parallel: on the one hand, an Adequacy Reviewer-Agent that verifies the initial translation for terminological and factual adequacy, and provides adequacy improvement suggestions, if applicable; on the other hand, a Fluency Reviewer-Agent that evaluates the translation’s readability, clarity, and coherence, and provides fluency improvement suggestions, if applicable. Finally, an Editor-Agent, which oversees the integration of the reviewers’ outputs, resolves conflicts between adequacy and fluency suggestions, and ensures overall consistency. The instructions of the different AI agents are provided in Appendix \ref{sec:appendix}. A public demo of the system, which can be used with different language combinations, language models and files, is available at the following link: \href{https://agents-parallel-2.streamlit.app/}{https://agents-parallel-2.streamlit.app/.}

\subsection{The underlying MT systems}
To systematically assess the impact of the proposed multi-agent workflows, we compare six system configurations: four multi-agent workflows with different model temperatures and two state-of-the-art NMT systems:

\begin{itemize}
    \item \textbf{Multi-Agent Big 1.3}: In this configuration, all AI agents use DeepSeek R1 (671B parameters) with a temperature setting of 1.3 \citep{deepseek-aiDeepSeekR1IncentivizingReasoning2025}. This choice balances creativity and precision, ensuring that the system can generate fluent yet legally precise translations while allowing flexibility in phrasing when necessary. With both "Multi-Agent Big" workflows, we aim to assess how big LLMs behave in multi-agent MT systems.
    \item \textbf{Multi-Agent Big 1.3/0.5:} This configuration also employs DeepSeek R1 but introduces a differentiated temperature setting strategy. The Translator-Agent and Editor-Agent operate at a temperature of 1.3, promoting creative phrasing where appropriate. Meanwhile, the Adequacy Reviewer-Agent and Fluency Reviewer-Agent function at a temperature of 0.5, prioritizing deterministic validation and reducing variability in term consistency and grammatical precision.
    \item \textbf{Multi-Agent Small 1.3:} In this configuration, all AI agents utilize gpt-4o-mini-2024-07-18 (unknown parameters, but reportedly a smaller language model) with a temperature setting of 1.3 for every agent. With both "Multi-Agent Small" workflows, we aim to assess how small LLMs behave in multi-agent MT systems.
    \item \textbf{Multi-Agent Small 1.3/0.5:} This configuration also employs gpt-4o-mini-2024-07-18 but introduces a differentiated temperature setting. The Translator-Agent and Editor-Agent operate at a temperature of 1.3, while the Adequacy Reviewer-Agent and Fluency Reviewer-Agent function at a temperature of 0.5.
    \item \textbf{DeepL:} The baseline comparison consists of two widely used NMT systems. First, DeepL. As there are two model options (“Next-gen language model” and “Classic language model”) and we wanted to assess NMT, we opted for the “Classic language model” option.
    \item \textbf{Google Translate:} The second NMT system is Google Translate. These two NMT systems represent the current industry standard for MT and are among the most widely used worldwide.
\end{itemize}

\subsection{Document and evaluation}
A legal contract, originally written in English, serves as the test document. The text contains 2547 words, 100 segments, and a type-token ratio of 0.27, demonstrating a complex document pertaining to the legal domain. It includes several problematic elements, such as numbers and currencies, in-domain terminology, and complex structures. Therefore, it is a high-stakes, domain-specific document where terminological adequacy, syntactic structure, and legal compliance are critical. These complexities were chosen to see how the different MT system configurations would behave.

\begin{figure*}[t]
  \centering
  \includegraphics[width=\textwidth]{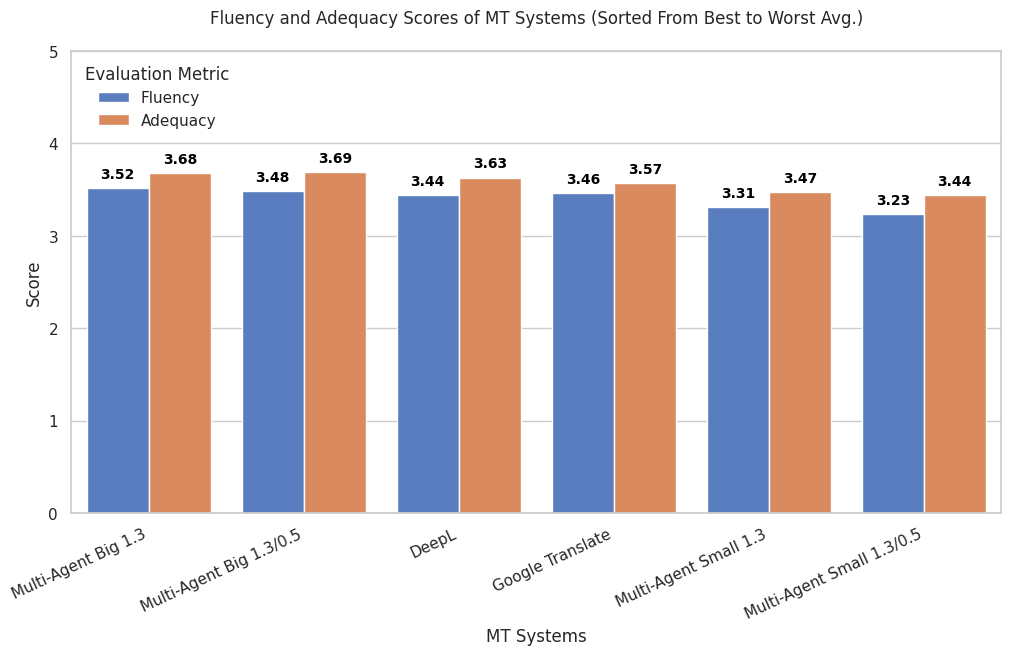}
  \caption{Fluency-Adequacy results.}
  \label{fig:adequacy-fluency}
\end{figure*}

A professional translator with +10 years of experience was recruited to evaluate the different MT outputs by following best practices for human evaluation of translation quality \citep{laubliSetRecommendationsAssessing2020}. Strict evaluation guidelines were provided (following the methodology in \citet{briva-iglesiasImpactTraditionalInteractive2023}). The complete data set is shared in \href{https://zenodo.org/records/15235526}{Zenodo}. The evaluator assessed a total of 15,282 words via different dimensions, namely:

\textbf{Adequacy:} The evaluator had to verify whether the translation preserved the meaning of the source text, including legal terminology, factual correctness, and adherence to jurisdictional requirements. On a scale from 1 (the lowest adequacy) to 4 (the highest adequacy).

\textbf{Fluency:} The evaluator had to assess readability, naturalness, and linguistic coherence, ensuring that the translation was stylistically appropriate for professional legal communication. On a scale from 1 (the lowest fluency) to 4 (the highest fluency).

\textbf{Ranking:} The evaluator compared the multiple MT outputs for the same source text and ranked from best (ranking score 1) to worst (ranking score 6). Instead of assigning absolute scores, the evaluator determined which translation was the best, second-best, and so on; ties were allowed.

\section{Discussion of the results}
This section presents the results of the comparative evaluation of the multi-agent and the NMT systems. First, the fluency and adequacy scores are discussed, followed by the overall ranking distribution.

\begin{figure*}[t]
  \centering
  \includegraphics[width=\textwidth]{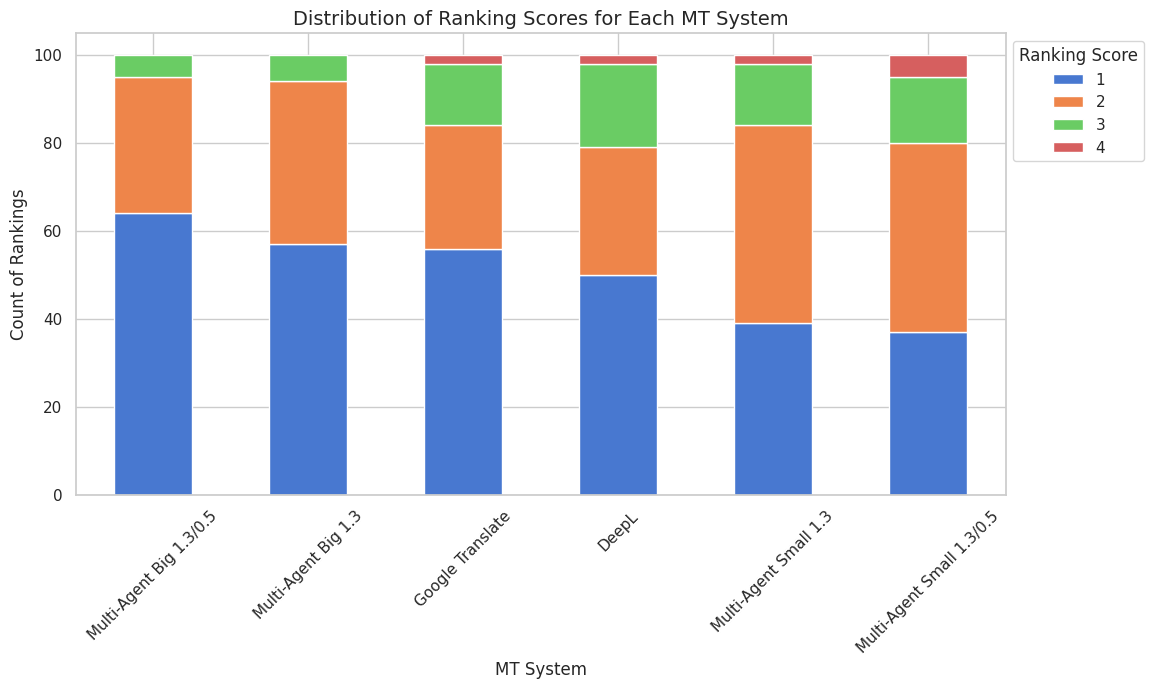}
  \caption{Ranking results.}
  \label{fig:ranking}
\end{figure*}

Figure \ref{fig:adequacy-fluency} presents the average fluency and adequacy scores across the six MT configurations analysed, sorted from highest to lowest based on their combined average scores. The two best-performing systems—Multi-Agent Big 1.3 and Multi-Agent Big 1.3/0.5—achieved similar results, with minor variations. Multi-Agent Big 1.3 ranked highest in fluency (3.52) and obtained an adequacy score of 3.68, making it the strongest individual system overall. Multi-Agent Big 1.3/0.5, on the other hand, achieved the highest adequacy score (3.69) while maintaining strong fluency (3.48), suggesting that the Multi-Agent Big approach obtained better results than state-of-the-art NMT systems. 

Both NMT systems analysed, DeepL and Google Translate, obtained lower scores than both Multi-Agent Big configurations, but higher scores than the Multi-Agent Small systems. The two worst performing systems—Multi-Agent Small 1.3 and Multi-Agent Small 1.3/0.5—consistently scored lower across both fluency and adequacy metrics. Multi-Agent Small 1.3 had a fluency score of 3.31 and an adequacy score of 3.47, slightly outperforming Multi-Agent Small 1.3/0.5, which scored 3.23 in fluency and 3.44 in adequacy. These scores positioned the multi-agent workflows powered by smaller LLMs at the lower end of the performance spectrum.
 
To complement the fluency and adequacy metrics, a ranking-based evaluation was conducted, where the evaluator assigned ordinal rankings (1st to 6th place) to each system's translations. Since there were some ties in every segment, ranking scores only range from 1 to 4. Figure \ref{fig:ranking} reveals that Multi-Agent Big 1.3 secured the highest proportion of first-place rankings (64 out of 100), followed closely by the Multi-Agent Big 1.3/0.5 system (57 out of 100). DeepL, despite its higher average score overall within the NMT systems, received fewer first-place rankings (50) than Google Translate (56), indicating that while it produced adequate outputs, it may have struggled with certain domain-specific fluency constraints. By conducting a more qualitative analysis, we can see that the English text “USD 1,000,000” was incorrectly translated into Spanish by the NMT systems as "\$1.000.000” (DeepL) and "USD 1,000,000" (Google Translate). In Spanish, the dollar sign should go after the number, and dots should be the thousands separator. All the multi-agent systems (both in Big and Small sizes) correctly translated this currency as “1.000.000 USD”. Similarly, Multi-Agent Systems demonstrated higher contextual coherence than NMT systems. The term "Agreement" was coherently translated by all the multi-agent systems as "Acuerdo" or "Convenio", while the NMT systems offered different translations for the same source concept within the same translation. 

The two best performing systems were scored only with the ranking scores 1, 2 and 3, while both NMT systems and the Multi-Agent Small configurations had a modest presence in the ranking scores 3 and 4. The Multi-Agent Small 1.3 and Multi-Agent Small 1.3/0.5 systems were rated with score 1 in only 39 and 37 instances, respectively, further reinforcing the observation that model size significantly impacts translation performance in a multi-agent setup.

Given the modest evaluation size and language pairs used in this study, definitive conclusions cannot yet be drawn. However, the findings provide valuable insights into the potential of multi-agent systems for MT and suggest several promising avenues for future research. The results indicate that multi-agent workflows may obtain higher translation quality than NMT systems. Both Multi-Agent Big configurations outperformed traditional NMT models in adequacy and fluency. This suggests that integrating multiple specialized agents into MT workflows may allow for greater domain adaptation and content preservation, particularly in high-stakes fields such as legal and medical translation.

Despite these promising results, the current multi-agent system was implemented with no external tools. The inclusion of memory, RAG, domain- and client-specific databases, and more granular agent role customization could further improve performance \citep{liSurveyRetrievalAugmentedText2022}. Future studies should explore how additional tooling and fine-tuned role assignments influence translation quality in multi-agent systems.

The study also suggests that the temperature setting plays a significant role in MT outcomes. Higher temperatures for Reviewer-Agents correlated with stronger adequacy and fluency scores. A systematic investigation into the optimal balance between creative (higher temperature) and deterministic (lower temperature) agent behaviours could provide deeper insights into best practices for multi-agent MT workflows. The results also demonstrate that larger models tend to perform better in multi-agent settings. The Multi-Agent Big configurations consistently outperformed the smaller Multi-Agent Small systems, indicating that computational capacity is a critical factor in achieving high translation quality. Future work should examine the trade-offs between computational efficiency and translation quality, particularly for organizations with limited resources.

\section{Conclusion}
This paper has provided one of the early analysis of multi-agent systems in MT, comparing their performance with traditional NMT. First, we provided a thorough overview of the potential of AI agents for MT, both from a theoretical perspective—by exploring different workflows, potential use cases, and system architectures—as well as from a practical perspective—through a modest pilot study and a public demo designed for replication and further analysis. 

The paper opens an entirely new area of research focused on identifying optimal multi-agent configurations for MT and enhancing multilingual digital communication. Our findings highlight that multi-agent workflows obtain higher translation quality than traditional NMT systems and/or single-agent systems in our specific use case. Research on multi-agent systems is still in the early stages, and substantial empirical research is needed. So far, our pilot study highlights the impact of model size and temperature tuning on translation performance. Besides these key findings, several areas for future research emerge:

\textbf{Tool integration:} What is the impact of integrating external resources such as RAG, translation memories, specialized glossaries, and legal/medical databases to different multi-agent workflows? Also, what agent should acquire this knowledge? It is worth stressing that our multi-agent system is a basic workflow that obtains great results without tool access. Adding tool access is a simple task that may improve translation performance even further, if compared with NMT, which would need to be fine-tuned to acquire this specific knowledge. 

\textbf{Scaling multi-agent systems:} The scalability of multi-agent workflows for larger datasets and broader language pairs needs to be addressed. What is the performance of LLM-powered multi-agent systems in minor languages?

\textbf{Evaluation methodologies:} Developing more rigorous human and automated evaluation frameworks tailored to multi-agent MT workflows is required, as the potential workflows are limitless. How can we ensure a reliable evaluation of multi-agent systems?

\textbf{Cost and resource optimization:} Exploring the trade-offs between performance and sustainability, including token usage, computational costs, and energy efficiency in large-scale translation operations, is a crucial next step. Resource optimization, particularly token management, is a critical factor. The cost of computing power and tokens includes all inputs fed to the translator, reviewer, and editor agents. While language model costs are decreasing, sustainability remains a pressing issue. One promising avenue is to explore hybrid workflows where low-cost models handle simple tasks while high-performance models are reserved for complex texts, ensuring both cost-effectiveness and sustainability.

\textbf{Human-AI collaboration:} Examining how MT users interact with AI agents in translation workflows is also of relevance, not only in professional translation, but also for MT users beyond the language services industry, as most MT users are not professional translators. How can multi-agent systems be used for bridging language barriers and enhance multilingual digital communication in a human-centered way? \citep{briva-iglesiasFosteringHumancenteredAugmented2024a}

This said, AI agents may represent a new frontier in MT, offering dynamic solutions to the rigidity of traditional MT systems. By integrating autonomy, context-awareness, and iterative refinement, multi-agent systems may be able to enhance translation quality, scalability, and adaptability across domains. Yet, this is still to be empirically tested.

Beyond technical advancements, AI agents unlock opportunities for societal equity, from bridging language divides in education and crisis response to preserving endangered linguistic heritage. However, their deployment is not without challenges. Technical hurdles like latency and model dependency, ethical concerns around bias and accountability, and economic barriers such as high development costs demand urgent attention.

The path forward requires interdisciplinary collaboration, ethical stewardship, and sustainable innovation. Researchers must prioritize robust evaluation frameworks and low-resource language support, while industry stakeholders should invest in human-centered designs and green AI infrastructure. Translators, as critical partners, will also need upskilling to navigate hybrid workflows that blend human expertise with AI efficiency.

\bibliography{mtsummit25}

\appendix

\section{Appendix}
\label{sec:appendix}

This Appendix contains the Roles of the different AI agents of the multi-agent system analysed in the paper. The system instructions and the code are not shared due to it being a proprietary product.

\begin{table*}[h]
    \centering
    \begin{tabular}{|p{4cm}|p{12cm}|}
        \hline
        \textbf{Role} & \textbf{Description} \\
        \hline
        \textbf{Translator-Agent} & 
        You are a senior legal translator specializing in Intellectual Property documents. \newline
        Translate the provided legal text from \textbf{[source language]} to \textbf{[target language]} with perfect accuracy, legal terminology consistency, and publication-ready quality. \newline
        Return ONLY the translation with no additional text, spaces, or commentary. \\
        \hline
        \textbf{Adequacy Reviewer-Agent} & 
        You are an Adequacy Reviewer specializing in \textbf{[source language]} to \textbf{[target language]} translations. \newline
        \textbf{Strict instructions:} Review the current translation for adequacy issues (such as mistranslations, omissions, or untranslated segments) and output only a list of suggestions as plain text bullet points. \newline
        Maintain original style and format. \newline
        Each suggestion must be formatted exactly as: \texttt{ERROR: [issue] → SUGGESTION: [fix]}. \newline
        If no corrections are needed, output \textbf{"Accuracy: No corrections needed"}. \newline
        Do not include any additional text, commentary, or the corrected translation—only the bullet-point list of suggestions. \\
        \hline
        \textbf{Fluency Reviewer-Agent} & 
        You are a Fluency Reviewer specializing in \textbf{[source language]} to \textbf{[target language]} translations. \newline
        \textbf{Strict instructions:} Review the current translation for fluency issues (including grammar, spelling, natural flow, and cultural adaptation) and output only a list of suggestions as plain text bullet points. \newline
        Focus only on: Grammar/spelling errors; Natural flow in \textbf{[target language]}; Cultural adaptation. \newline
        Each suggestion must be formatted exactly as: \texttt{ERROR: [issue] → SUGGESTION: [fix]}. \newline
        If no corrections are needed, output \textbf{"Fluency: No corrections needed"}. \newline
        Do not include any additional text or commentary—only the bullet-point list of suggestions. \\
        \hline
        \textbf{Editor-Agent} & 
        You are a senior legal editor specializing in legal documents. Your task is to integrate the first translation with the accuracy and fluency suggestions to produce the final polished translation. \newline
        \textbf{Strict instructions:} Output only the final translation as a single plain text string with no additional commentary, labels, or formatting. \newline
        Maintain legal accuracy and preserve the document’s technical structure. \\
        \hline
    \end{tabular}
    \caption{AI Agent Instructions}
    \label{tab:roles}
\end{table*}

\end{document}